\documentclass[a4paper,fleqn]{cas-sc}
\usepackage{amsmath,amssymb,booktabs,threeparttable,multirow,rotating}
\usepackage{graphicx}
\usepackage{natbib}
\usepackage{capt-of}
\usepackage{xcolor}
\usepackage{float}
\usepackage{placeins}

\newsavebox{\weibotablebox}
\newsavebox{\weibofigurebox}
\newsavebox{\weibotablecapbox}
\newsavebox{\weibofigcapbox}

\newcommand{\msd}[2]{#1{\tiny$\pm$#2}}
\newcommand{\best}[2]{\textcolor{red}{#1{\tiny$\pm$#2}}}
\newcommand{\second}[2]{\textcolor{blue}{#1{\tiny$\pm$#2}}}

\definecolor{citeblue}{RGB}{0,102,153}

\hypersetup{
    colorlinks=true,
    citecolor=citeblue,
    linkcolor=black,
    urlcolor=citeblue
}

\graphicspath{{../}}

\begin{document}
\let\WriteBookmarks\relax
\def\floatpagepagefraction{1}
\def\textpagefraction{.001}

\shorttitle{Rethinking Multimodal Fake News Detection in the Generative AI Era}
\shortauthors{Wenbin Shen et~al.}

\title[mode=title]{Rethinking Multimodal Fake News Detection in the Generative AI Era}

\author[1]{Wenbin Shen}[
    orcid=0009-0008-3456-4003
]
\cormark[1]
\ead{shenwenbin@njust.edu.cn}

\author[1]{Guoxuan Qin}[
    orcid=0009-0001-1711-6443
]

\author[1]{Guangxu Yao}[
    orcid=0009-0003-9267-8838
]

\author[1]{Baodong Wang}[
    orcid=0009-0001-1381-7680
]

\author[1]{Yuanbo Rui}[
    orcid=0009-0005-3427-0679
]

\author[1,2]{Zhichao Lian}[
    orcid=0000-0002-6643-8975
]
\cormark[1]
\ead{lzcts@163.com}

\cortext[1]{Corresponding author}

\affiliation[1]{
  organization={Nanjing University of Science and Technology},
  city={Nanjing},
  country={China}
}

\affiliation[2]{
  organization={University of Chinese Academy of Sciences},
  city={Beijing},
  country={China}
}

\begin{abstract}
Generative content is increasingly entering the production and dissemination of news, transforming fake news from manually fabricated or simply manipulated material into complex forms in which native and generated content jointly participate. Existing multimodal fake news detection research primarily focuses on veracity assessment and rarely characterizes how generativity differences affect the reliability of evidence. In contrast, AIGC detection primarily determines whether content is generated or modified by generative models, but it does not by itself establish whether the underlying news event is true. To bridge the separation between these tasks in data and evaluation, we construct Weibo26, a multimodal fake news detection dataset for generative-content scenarios. On this basis, we propose the Generativity-Aware Hierarchical Reasoning (GAHR) framework, which combines global judgment with local correction so that generativity information participates in news-veracity reasoning. Experiments on multiple existing fake news detection benchmarks and Weibo26 show that GAHR achieves competitive veracity-detection performance while effectively identifying generative content.
\end{abstract}


\maketitle

\section{Introduction}

Social media has become an important channel for obtaining, publishing, and redistributing news, as well as a major medium through which fake news spreads \citep{1,2,3}. On social media, news is commonly reposted, rewritten, and recombined at speed. Assessing news veracity depends not only on an individual piece of content but also on whether different modalities jointly support the same event facts \citep{4,5}. In recent years, generative artificial intelligence has entered the production and dissemination of news on a broad scale. Large language models can produce coherent, fluent, news-style text \citep{35,36,37}. Diffusion and text-to-visual models have also substantially lowered the barrier to producing high-quality visual content \citep{39,40,41,42}. Consequently, fake news is no longer limited to manual fabrication, image--text mismatch, or local manipulation; it increasingly exhibits complex structures involving both native and generated content.

Multimodal fake news detection centers on news-veracity assessment and has developed a relatively rich body of datasets and methods \citep{6,7}. Relevant methods have progressed from modeling text, visual, and social-context signals to cross-modal representation learning, attention-based fusion, and consistency modeling \citep{8,9,10,16}. Recent studies have further improved discrimination on complex samples through fine-grained collaborative detection and large vision--language models \citep{27,28}. AIGC detection, in contrast, mainly concerns whether content is generated by a model. Generated-text detection typically identifies generated content through statistical distributions and probability features \citep{43,44,45,46}. Generated-visual-content detection focuses on clues such as frequency-domain anomalies, texture traces, and model fingerprints \citep{54,55,56,57}. These two lines of work respectively address veracity and generativity, but the former usually does not record generative-model involvement, whereas the latter cannot directly determine whether a news event is true.

\begin{figure}[pos=htbp]
  \centering
  \includegraphics[width=\textwidth]{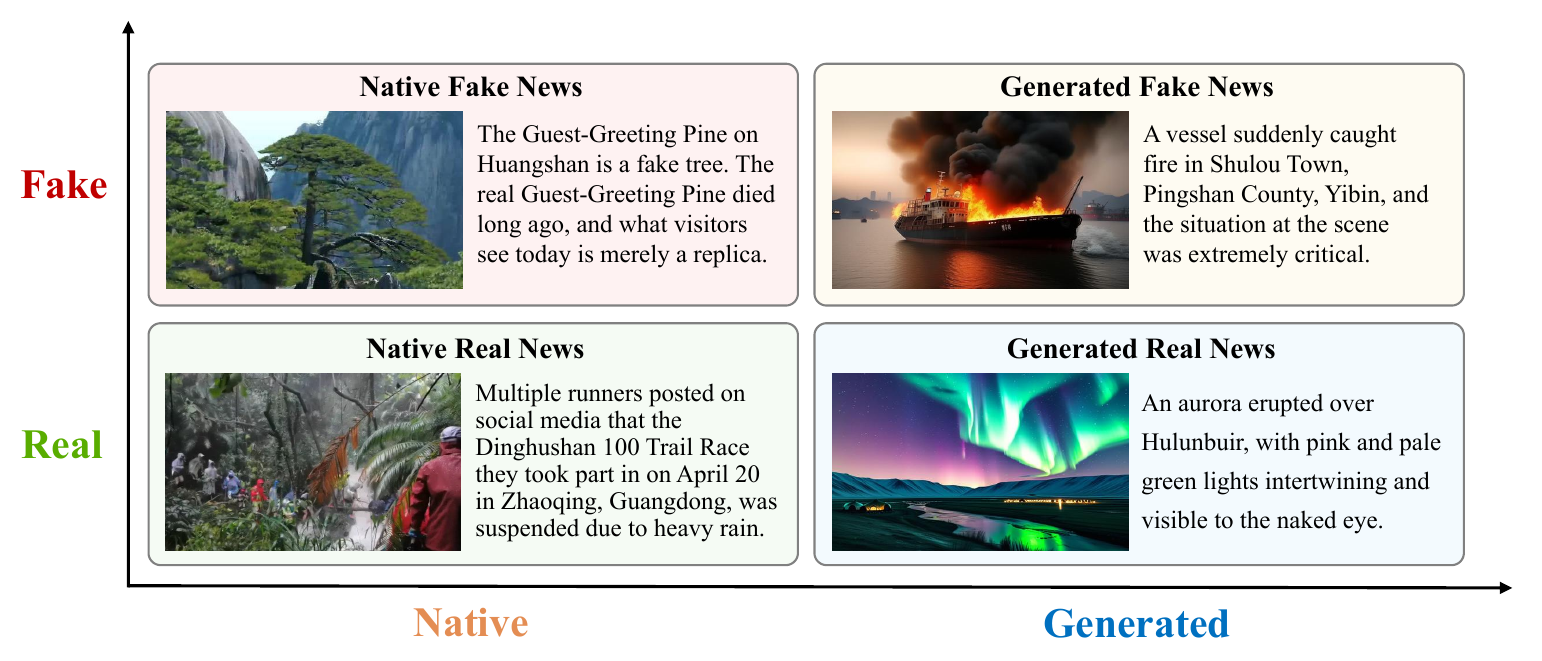}
  \caption{Conceptual organization of multimodal news samples along two distinct dimensions: veracity (real versus fake) and generative involvement (native versus generated). Native samples contain no generated content, whereas generated-content samples contain AI-generated content.}
  \label{fig:generative-era}
\end{figure}

In the generative-AI era, multimodal fake news detection faces new challenges at both the data and method levels. At the data level, conventional fake news detection datasets lack generativity information and therefore cannot support unified evaluation of news in which generative content participates in dissemination \citep{11,12}. At the method level, existing multimodal fusion models can model overall semantic associations but do not necessarily distinguish the relationship between generativity differences and veracity evidence; existing generated-content detectors can identify some generative traces but cannot replace news-veracity reasoning. Research indicates that generative models are changing the way misinformation is created and the boundaries of detection \citep{73,74}. Figure~\ref{fig:generative-era} illustrates the relationship between veracity detection and generativity detection in multimodal news analysis.

To bridge this dataset-level gap, we construct Weibo26, a multimodal fake news detection dataset for generative-content scenarios. The dataset is based on social-media news events and uses two text generation models and six image generation models, ultimately forming six sample types and containing 15,154 samples. Weibo26 provides both veracity and generativity labels for every sample. Compared with conventional multimodal fake news detection datasets, Weibo26 characterizes veracity differences under distinct types of generative involvement. Compared with standard generated-content detection datasets, it places generativity discrimination in the context of news-veracity assessment, so model evaluation does not stop at the question of ``whether it is generated.''

Based on Weibo26, we propose the Generativity-Aware Hierarchical Reasoning (GAHR) framework. GAHR first maps text, image, and cross-modal pre-trained features into a unified space and forms a global veracity judgment through bidirectional cross-modal interaction. It then extracts local evidence from the text side, image side, and image--text consistency, and corrects the global judgment in a gated manner. On data with generativity annotations, GAHR can further learn text- and image-generativity judgments, allowing generativity information to participate in training as an auxiliary signal.

The core contributions of this work are as follows.
\begin{enumerate}
  \item We construct Weibo26, a multimodal fake news detection dataset for generative-content scenarios. The dataset contains six sample types and provides both veracity and generativity labels.
  \item We propose the Generativity-Aware Hierarchical Reasoning (GAHR) framework. It jointly models global judgment and local correction, enabling generativity information to participate in veracity reasoning.
  \item We conduct systematic experiments on five existing multimodal fake news detection benchmarks and Weibo26. The experiments validate the effectiveness of the proposed dataset and method from multiple perspectives, including veracity detection and AIGC detection.
\end{enumerate}

\section{Related Work}
\subsection{Multimodal Fake News Detection}

Multimodal fake news detection aims to assess news veracity by exploiting multiple information sources in news content \citep{6,7}. Early studies commonly treated fake news detection as a content-classification problem and mainly relied on text representations, visual representations, or social-context features \citep{8,9,10}. Datasets such as LIAR and FakeNewsNet provide foundational evaluation settings for fact checking and social-media fake news detection \citep{11,12,13,14}. As image--text news dissemination has become commonplace, a single modality may not provide sufficient evidence for assessing multimodal news content. Jin et al. were among the early researchers to apply multimodal fusion to Weibo rumor detection \citep{15}. EANN reduces event bias through event-adversarial learning, MVAE learns joint multimodal representations, and the SpotFake series enhances image--text feature representations through transfer learning \citep{16,17,18,19}.

Existing methods mainly focus on representation fusion, cross-modal relation modeling, knowledge enhancement, and uncertainty-aware reasoning. Representation-fusion methods generally first extract semantic features from different modalities and then obtain a joint representation using concatenation, weighting, gating, or Transformer architectures \citep{17,18,21,30}. Similarity- and inconsistency-modeling methods explicitly characterize the relationship between modalities, including whether textual and visual contents are semantically aligned or mutually inconsistent \citep{20,22}. Knowledge- and entity-enhanced methods further use external knowledge, vision-language models, or entity cues to supplement content evidence \citep{24,25,29}. More recent studies improve detection from the perspectives of uncertainty modeling, fine-grained collaborative detection, and multi-view large vision-language modeling \citep{23,26,27,28,38}. Overall, existing datasets and evaluation protocols primarily serve real--fake classification in conventional dissemination environments and are difficult to use to fully cover changes in news sample forms and evidence structures after generative content becomes involved \citep{31,32,33,34}.

\subsection{AIGC Detection}

AIGC detection primarily determines whether content is produced by generative models or has been edited with their involvement. Generated-text detection typically exploits differences in lexical distributions, syntactic patterns, and probability distributions \citep{43,44,45,46}. Prior work has shown that detector effectiveness is closely related to the generation model, data domain, and post-processing approach \citep{47,52,53}. DetectGPT uses probability curvature for zero-shot detection, whereas watermarking methods attempt to embed verifiable signals during generation \citep{48,49}. As generative models continue to evolve, detectors that rely only on superficial statistical features may face difficulties when generalizing across models, domains, and post-processing conditions. Benchmarks such as HC3 and M4 further extend evaluation to multi-generator, multi-domain, and multilingual settings \citep{50,51}.

Generated-visual-content detection mainly focuses on frequency-domain anomalies, local textures, compression traces, and model fingerprints in generated images \citep{54,55,56,57,61}. Detection studies for diffusion models show that even high-quality generated images may retain exploitable structural differences \citep{58}. Recent methods such as Text-aware CLIP attempt to connect generated-visual-content detection and fake news detection through cross-modal pre-trained representations \citep{72}. These works suggest that recent AIGC detection studies increasingly consider semantic and structural cues in addition to low-level artifacts, but its data construction and evaluation objectives still mainly focus on generativity discrimination, making it difficult to support unified evaluation of news veracity, modes of generative involvement, and their interactions.

\section{Construction of the Weibo26 Dataset}

\subsection{Data Collection and Cleaning}
\label{sec:data-cleaning}

Weibo26 is constructed in two stages: the collection and cleaning of original news samples, followed by the construction and verification of derived samples. The original data are collected from publicly available social-media news, authoritative media reports, rumor-debunking records, and fact-checking sources. For real news, each sample must be verifiable through authoritative reports, official releases, or cross-validation by multiple reliable sources. For fake news, each sample must be supported by explicit debunking evidence or fact-checking records. Controversial opinions, satirical expressions, and content lacking stable factual evidence are excluded.

Initial cleaning is conducted on both the text and image modalities. On the text side, residual HTML tags, special characters, and garbled content are removed, followed by duplicate removal and sample screening. On the image side, samples with invalid links, corrupted files, or images that cannot be decoded normally are detected and discarded. Samples with missing text, missing images, incomplete image--text structures, or insufficient factual evidence are excluded from subsequent construction.

After cleaning, the retained original samples are screened according to the requirements of the corresponding construction task. For each original sample selected for derived-sample construction, a one-to-one mapping is established between the original sample and its derived sample. Original samples that are not selected for generative construction constitute the native NRN and NFN subsets. For each original sample selected for derived-sample construction, the source sample is used only as the factual and semantic basis for generation and is excluded from the final Weibo26 dataset. Therefore, the final dataset never contains both an original source sample and its derived counterpart from the same construction instance.

To ensure sample quality, Weibo26 undergoes an additional manual review after cleaning and preliminary construction. The review examines whether each sample has a complete content structure, whether it is sufficiently classifiable, and whether its veracity label can be supported by publicly available evidence. For generated samples, the reviewers additionally verify whether the generated content satisfies the intended generation condition and whether its relation to the source information is factually valid. A sample is retained only after agreement among all three reviewers. Rejected or ambiguous samples are either reconstructed under revised generation conditions or excluded, and any reconstructed samples undergo the same review procedure.

\subsection{Construction of Generative Samples}

Weibo26 constructs derived samples using the cleaned original news as the factual and semantic basis. The original news used in this process is not directly incorporated together with its corresponding derived sample into the final dataset. Instead, each selected original sample is used to construct one corresponding derived sample according to a predefined generation task, after which the generated result is subjected to manual verification before being retained.

Text generation uses DeepSeek-V4-Flash \citep{76} and Qwen3-VL \citep{77}. Image generation uses six models: Flux \citep{66}, Hunyuan \citep{67}, Kandinsky \citep{68}, Kolors \citep{69}, Pixart \citep{70}, and SDXL \citep{71}. Flux, Hunyuan, Kandinsky, and Kolors cover different diffusion-model implementations, while Pixart and SDXL supplement visual-generation samples based on DiT and latent-diffusion architectures.

The generated-text samples include two types. The first type generates factually consistent news text conditioned on real-news images and corresponds to ICRN. The generated text must preserve the key facts of the source news, including its principal entities, locations, time information, event relations, and outcomes. The second type generates news text containing factual deviations conditioned on real-news images and corresponds to ICFN. The introduced deviations mainly concern key entities, locations, times, quantities, causal relations, or event outcomes, while maintaining a complete and coherent news expression.

The generated-image samples also include two types. The first type generates factually consistent images conditioned on real-news text and corresponds to TCRN. The generated image must be semantically compatible with the source text and preserve the main event information. The second type generates images matching fake narratives conditioned on fake-news text and corresponds to TCFN. The generated image must be visually consistent with the corresponding false description and support the intended misleading narrative.

Generation is followed by a dedicated manual verification process rather than direct inclusion of all generated results. For ICRN and TCRN, the reviewers verify whether the generated content remains factually consistent with the corresponding real-news source. For ICFN and TCFN, they verify whether the intended factual deviation or misleading relation is clearly expressed and whether the generated modality is consistent with the corresponding false narrative. Only samples unanimously approved by all three reviewers are retained. Samples that fail verification are reconstructed by modifying the generation instructions or generation conditions and are reviewed again until they satisfy the predefined criteria.

\subsection{Analysis of the Weibo26 Dataset}
\label{sec:weibo26-analysis}

Weibo26 provides a veracity label and a generativity label for each sample. The veracity label indicates whether news content is consistent with verifiable facts, whereas the generativity label indicates whether a sample contains generative content. The two labels describe different properties: the veracity label concerns factual consistency, whereas the generativity label concerns whether the content is generated.

\begin{figure}[pos=h]
    \centering
    \includegraphics[width=\textwidth]{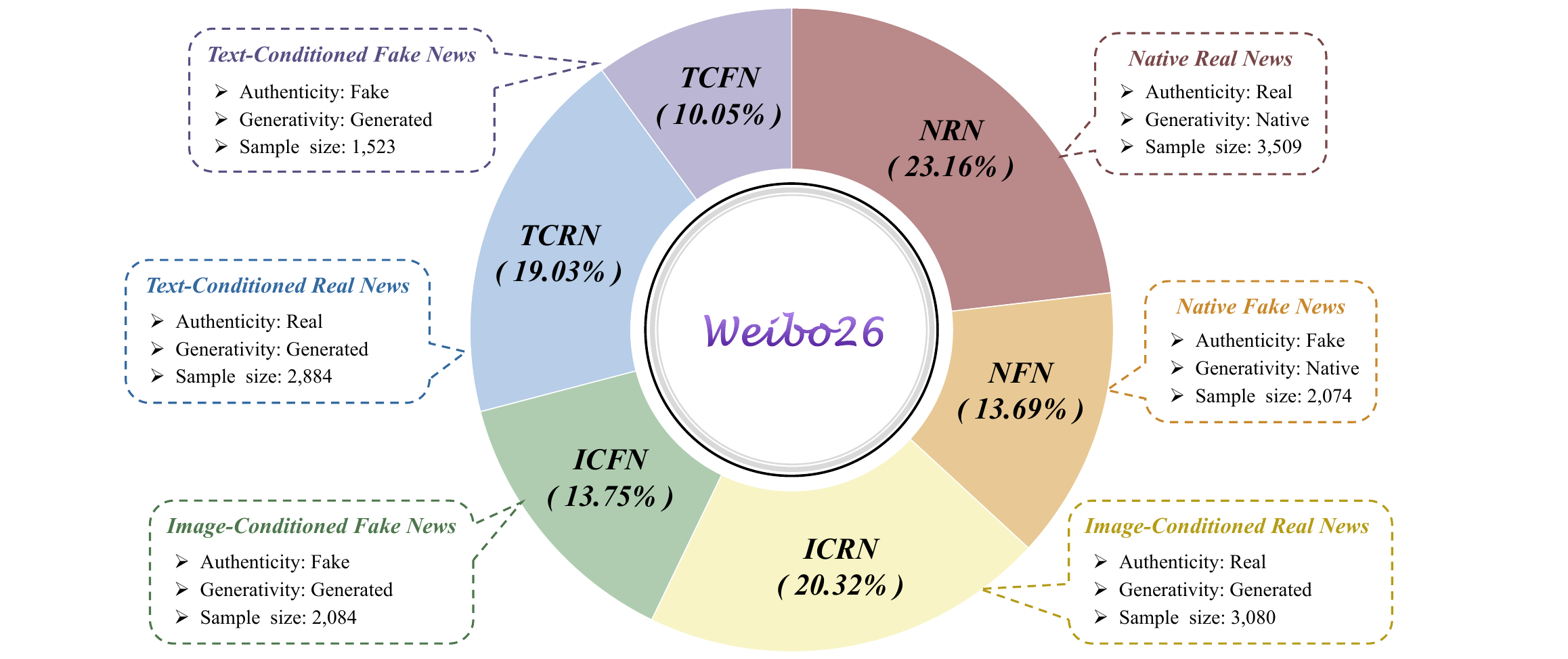}
    \caption{Overview of the six sample types in Weibo26, including their statistics and distribution.}
    \label{fig:weibo26-overview}
\end{figure}

Weibo26 contains 15,154 image--text samples divided into six types. NRN and NFN are native news samples, corresponding to real and fake news in the real world, respectively. ICRN and ICFN are image-conditioned text samples, respectively corresponding to real and fake text generated from real-news images. TCRN and TCFN are text-conditioned image samples, respectively corresponding to factually consistent images generated from real-news text and images generated from fake-news text that match fake narratives. Figure~\ref{fig:weibo26-overview} shows the statistics for the different sample types.

In terms of label distribution, real samples comprise NRN, ICRN, and TCRN, totaling 9,473 instances; fake samples comprise NFN, ICFN, and TCFN, totaling 5,681 instances. There are 5,583 native samples and 9,571 samples involving generative content. To further illustrate the difference between Weibo26 and existing multimodal fake news detection datasets, Table~\ref{tab:dataset-comparison} compares language, publication year, temporal range, and sample scale. Compared with existing datasets, Weibo26 retains news-veracity labels while introducing generative-sample construction, enabling model evaluation in dissemination scenarios involving generative content.

\begin{table*}[pos=h]
\caption{Comparison of Weibo26 with existing multimodal fake news detection datasets.}
\label{tab:dataset-comparison}
\centering
\begin{tabular}{llllrrr}
\toprule
Dataset & Language & Publication year & Time range & Real & Fake & Image\\
\midrule
PHEME \citep{75}      & English & 2017 & 2014--2015 & 3,830   & 1,972   & 5,802\\
GossipCop \citep{12}  & English & 2020 & up to 2020 & 16,817  & 5,323   & 18,417\\
Fakeddit \citep{33}   & English & 2020 & 2008--2019 & 527,049 & 628,501 & 682,996\\
AMG \citep{78}        & English & 2025 & 2016--2024 & 3,018   & 2,004   & 5,022\\
Weibo17 \citep{15}    & Chinese & 2017 & 2012--2016 & 4,779   & 4,749   & 9,528\\
Weibo21 \citep{5}     & Chinese & 2021 & 2014--2021 & 4,640   & 4,488   & 9,128\\
CFND \citep{79}       & Chinese & 2024 & up to 2024 & 16,394  & 10,271  & 26,665\\
\midrule
Weibo26              & Chinese & 2026 & 2024--2026 & 9,473   & 5,681   & 15,154\\
\bottomrule
\end{tabular}
\end{table*}

\section{Methodology}
\subsection{Overview}

\begin{figure}[pos=h]
  \centering
  \includegraphics[width=\textwidth]{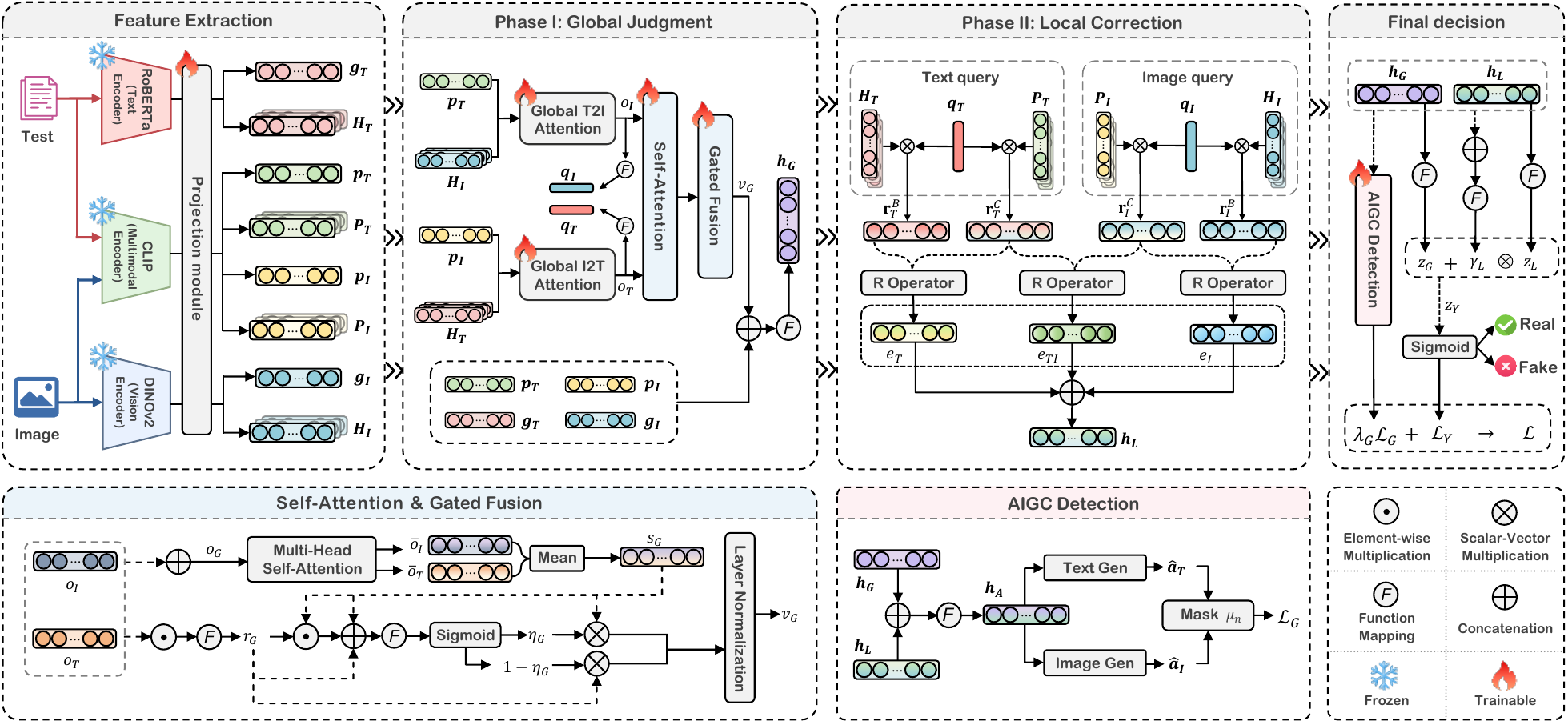}
  \caption{Overall architecture of the Generativity-Aware Hierarchical Reasoning (GAHR) framework.}
  \label{fig:gahr}
\end{figure}

The overall architecture of GAHR is shown in Fig.~\ref{fig:gahr}. The model first projects text, image, and cross-modal encoder outputs into a common representation space. At the global stage, bidirectional cross-modal attention then produces global representations for veracity prediction. At the local stage, directional responses guide the extraction of text-side, image-side, and image--text consistency evidence. A gated correction head combines this local evidence with the global judgment. When generativity labels are available, an auxiliary branch predicts text- and image-side generative-content status.

\subsection{Unified Encoding Representation}

Given a news sample, let its text and image be denoted by $\mathcal{T}$ and $\mathcal{I}$, respectively. GAHR uses a text encoder, a visual encoder, and a cross-modal pre-trained encoder to extract base semantic features and cross-modal alignment features. The base encoding process is
\begin{equation}
(\mathbf{x}_{T},\mathbf{X}_{T})=F_T(\mathcal{T}),\quad
(\mathbf{x}_{I},\mathbf{X}_{I})=F_I(\mathcal{I}).
\end{equation}
Here, $F_T$ and $F_I$ are the text and visual encoders; $\mathbf{x}_{T}$ and $\mathbf{x}_{I}$ are base global representations; and $\mathbf{X}_{T}$ and $\mathbf{X}_{I}$ are base local sequences. Lowercase bold variables denote global representations, uppercase bold variables denote local sequences, and the subscripts $T$ and $I$ denote the text and image sides, respectively.

The cross-modal pre-trained encoder produces another set of representations in an aligned space:
\begin{equation}
(\mathbf{c}_{T},\mathbf{C}_{T})=F_C^T(\mathcal{T}),\quad
(\mathbf{c}_{I},\mathbf{C}_{I})=F_C^I(\mathcal{I}).
\end{equation}
Here, $F_C^T$ and $F_C^I$ are the text-side and image-side encoders of the cross-modal pre-trained model; $\mathbf{c}_{T}$ and $\mathbf{c}_{I}$ are cross-modal global representations; and $\mathbf{C}_{T}$ and $\mathbf{C}_{I}$ are cross-modal local sequences.

Because encoder output dimensions are not fully consistent, GAHR maps every representation into a unified dimension $d$ by a projection function. For an arbitrary feature $\mathbf{a}$, it is defined as
\begin{equation}
\phi_r(\mathbf{a})=\operatorname{GELU}(\operatorname{LN}(\mathbf{a}\mathbf{W}_{r}+\boldsymbol{\delta}_{r})).
\end{equation}
Here, $\phi_r$ is the projection function for the $r$-th feature type, $\mathbf{W}_r$ and $\boldsymbol{\delta}_r$ are learnable parameters, $\operatorname{LN}$ is layer normalization, and $\operatorname{GELU}$ is a nonlinear activation function.

Let $M\in\{T,I\}$ denote the modality index. The projected base semantic features are
\begin{equation}
(\mathbf{g}_{M},\mathbf{H}_{M})=(\phi_{gM}(\mathbf{x}_{M}),\phi_{hM}(\mathbf{X}_{M})),\quad M\in\{T,I\}.
\end{equation}
Here, $\mathbf{g}_{T}$ and $\mathbf{g}_{I}$ are base global representations, while $\mathbf{H}_{T}$ and $\mathbf{H}_{I}$ are base local sequences.

Cross-modal alignment features are projected into the same dimensional space:
\begin{equation}
(\mathbf{p}_{M},\mathbf{P}_{M})=(\phi_{cM}(\mathbf{c}_{M}),\phi_{CM}(\mathbf{C}_{M})),\quad M\in\{T,I\}.
\end{equation}
Here, $\mathbf{p}_{T}$ and $\mathbf{p}_{I}$ are cross-modal global representations, and $\mathbf{P}_{T}$ and $\mathbf{P}_{I}$ are cross-modal local sequences. All subsequent modules operate in this unified dimensional space.

\subsection{Global Judgment}

The global judgment component forms an overall veracity assessment of a news image--text pair. GAHR first uses the cross-modal global representation of one modality to query the base local sequence of the other modality, producing cross-modal responses in two directions. Given a query vector $\mathbf{u}$, key sequence $\mathbf{K}$, value sequence $\mathbf{V}$, and position mask $\mathbf{m}$, the attention score at position $j$ is defined as
\begin{equation}
s_j=\frac{(\mathbf{W}_{q}^{a}\mathbf{u})^\top(\mathbf{W}_{k}^{a}\mathbf{K}_{j})}{\tau\sqrt{d}}.
\end{equation}
Here, $s_j$ is the attention score at position $j$, $\mathbf{W}_{q}^{a}$ and $\mathbf{W}_{k}^{a}$ are the projection matrices for queries and keys, and $\tau$ is the temperature coefficient.

Under the position-mask constraint, the attention weight is
\begin{equation}
\alpha_j=\frac{\exp(s_j)m_j}{\sum_{\ell=1}^{L_m}\exp(s_{\ell})m_{\ell}}.
\end{equation}
Here, $\alpha_j$ is the attention weight at position $j$, $m_j$ is the corresponding element of the position mask $\mathbf{m}$, and $L_m$ is the length of the queried sequence.

Based on these weights, the cross-modal attention function is
\begin{equation}
\operatorname{Att}(\mathbf{u},\mathbf{K},\mathbf{V},\mathbf{m})=\sum_{j=1}^{L_m}\alpha_j\mathbf{W}_{v}^{a}\mathbf{V}_j.
\end{equation}
Here, $\operatorname{Att}$ is the cross-modal attention function, $\mathbf{W}_{v}^{a}$ is the value projection matrix, $\mathbf{u}\in\mathbb{R}^{d}$, $\mathbf{K}\in\mathbb{R}^{L_m\times d}$, $\mathbf{V}\in\mathbb{R}^{L_m\times d}$, and $\mathbf{m}\in\{0,1\}^{L_m}$.

Accordingly, the model obtains an image response guided by text and a text response guided by images:
\begin{equation}
\mathbf{o}_{I}=\operatorname{Att}(\mathbf{p}_{T},\mathbf{H}_{I},\mathbf{H}_{I},\mathbf{1}_{I}),\quad
\mathbf{o}_{T}=\operatorname{Att}(\mathbf{p}_{I},\mathbf{H}_{T},\mathbf{H}_{T},\mathbf{m}_{T}).
\end{equation}
Here, $\mathbf{o}_{I}$ is the text-guided image response, $\mathbf{o}_{T}$ is the image-guided text response, $\mathbf{1}_{I}$ is an all-one mask for the image local sequence, and $\mathbf{m}_{T}$ is the valid-position mask for the text local sequence.

The two directional responses reflect how each modality selects content from the other. To model their relationship, GAHR organizes them as a bidirectional response sequence:
\begin{equation}
\mathbf{O}_{G}=[\mathbf{o}_{T};\mathbf{o}_{I}].
\end{equation}
Here, $\mathbf{O}_{G}\in\mathbb{R}^{2\times d}$ is the bidirectional response sequence, and the semicolon denotes stacking along the sequence dimension.

On this sequence, the output of the $h$-th self-attention head is
\begin{equation}
\mathbf{U}_{h}=\operatorname{softmax}\left(\frac{\mathbf{O}_{G}\mathbf{W}_{h}^{Q}(\mathbf{O}_{G}\mathbf{W}_{h}^{K})^{\top}}{\sqrt{d_h}}\right)\mathbf{O}_{G}\mathbf{W}_{h}^{V}.
\end{equation}
Here, $\mathbf{U}_h$ is the output of the $h$-th attention head, $d_h$ is the single-head dimension, and $\mathbf{W}_{h}^{Q}$, $\mathbf{W}_{h}^{K}$, and $\mathbf{W}_{h}^{V}$ are the projection matrices of that head.

The multi-head self-attention function is obtained by concatenating the outputs of all attention heads and applying a linear mapping:
\begin{equation}
\operatorname{MHA}(\mathbf{O}_{G})=\operatorname{Concat}(\mathbf{U}_{1},\ldots,\mathbf{U}_{H})\mathbf{W}^{O}.
\end{equation}
Here, $\operatorname{MHA}$ is the multi-head self-attention function, $H$ is the number of attention heads, and $\mathbf{W}^{O}$ is the output projection matrix.

The two positions in the multi-head self-attention output serve as the updated text and image responses:
\begin{equation}
[\bar{\mathbf{o}}_{T};\bar{\mathbf{o}}_{I}]=\operatorname{MHA}(\mathbf{O}_{G}).
\end{equation}
Here, $\bar{\mathbf{o}}_{T}$ and $\bar{\mathbf{o}}_{I}$ are the updated text and image responses, respectively.

Global interaction considers both the mean representation after self-attention and the multiplicative interaction of the original bidirectional responses. The former captures the combination of the two directional responses, whereas the latter preserves elementwise correspondence:
\begin{equation}
\mathbf{s}_{G}=\frac{1}{2}(\bar{\mathbf{o}}_{T}+\bar{\mathbf{o}}_{I}).
\end{equation}
Here, $\mathbf{s}_{G}$ is the global response representation after self-attention interaction.

The elementwise interaction of the original bidirectional responses is
\begin{equation}
\mathbf{r}_{G}=\phi_{m}(\mathbf{o}_{T}\odot\mathbf{o}_{I}).
\end{equation}
Here, $\mathbf{r}_{G}$ is the multiplicative interaction representation, $\phi_m$ is the corresponding projection function, and $\odot$ denotes elementwise multiplication.

The model then learns a fusion gate to integrate the two kinds of global interaction information:
\begin{equation}
\boldsymbol{\eta}_{G}=\sigma(f_{\eta}([\mathbf{s}_{G},\mathbf{r}_{G},\mathbf{s}_{G}\odot\mathbf{r}_{G}])).
\end{equation}
Here, $\boldsymbol{\eta}_{G}$ is the global fusion gate, $f_{\eta}$ is a gating mapping function, and $\sigma$ is the sigmoid function.

The fused global interaction representation is
\begin{equation}
\mathbf{v}_{G}=\operatorname{LN}(\boldsymbol{\eta}_{G}\odot\mathbf{s}_{G}+(\mathbf{1}-\boldsymbol{\eta}_{G})\odot\mathbf{r}_{G}).
\end{equation}
Here, $\mathbf{v}_{G}$ is the global interaction representation, $\mathbf{1}$ is an all-one vector, and $\operatorname{LN}$ is layer normalization.

The global representation consists of the cross-modal interaction representation, base global representations, and cross-modal global representations:
\begin{equation}
\mathbf{h}_{G}=f_{G}([\mathbf{v}_{G},\mathbf{g}_{T},\mathbf{g}_{I},\mathbf{p}_{T},\mathbf{p}_{I}]).
\end{equation}
Here, $\mathbf{h}_{G}$ is the global representation for veracity judgment, $f_G$ is the global mapping function, and $[\cdot]$ denotes vector concatenation.

The global classification head outputs the global veracity logit:
\begin{equation}
z_{G}=\mathbf{w}_{G}^{\top}\mathbf{h}_{G}+b_{G}.
\end{equation}
Here, $z_G$ is the global veracity logit, and $\mathbf{w}_{G}$ and $b_G$ are the weight vector and bias scalar of the global classification head. $z_G$ provides the overall judgment, which is subsequently adjusted by local correction.

\subsection{Local Correction}

The local correction component supplements detailed evidence that may be weakened in the global judgment. GAHR generates text-side and image-side queries from the bidirectional cross-modal responses:
\begin{equation}
\mathbf{q}_{T}=\mathbf{W}_{T}^{q}\mathbf{o}_{T}+\mathbf{b}_{T}^{q},\quad
\mathbf{q}_{I}=\mathbf{W}_{I}^{q}\mathbf{o}_{I}+\mathbf{b}_{I}^{q}.
\end{equation}
Here, $\mathbf{q}_{T}$ and $\mathbf{q}_{I}$ are the text-side and image-side local queries, and $\mathbf{W}_{T}^{q}$, $\mathbf{W}_{I}^{q}$, $\mathbf{b}_{T}^{q}$, and $\mathbf{b}_{I}^{q}$ are learnable parameters.

Given a local sequence $\mathbf{Z}$, a query vector $\mathbf{q}$, and a position mask $\mathbf{n}$, query-guided pooling first calculates position weights:
\begin{equation}
\omega_j=\frac{\exp(\mathbf{Z}_j^\top\mathbf{q}/\sqrt{d})n_j}{\sum_{\ell=1}^{L_z}\exp(\mathbf{Z}_{\ell}^\top\mathbf{q}/\sqrt{d})n_{\ell}}.
\end{equation}
Here, $\omega_j$ is the pooling weight at position $j$, $L_z$ is the local sequence length, and $n_j$ is the corresponding mask element.

The query-guided aggregation of the local sequence is defined as
\begin{equation}
\operatorname{Pool}(\mathbf{Z},\mathbf{q},\mathbf{n})=\sum_{j=1}^{L_z}\omega_j\mathbf{Z}_j.
\end{equation}
Here, $\operatorname{Pool}$ is the query-guided pooling function, and $\mathbf{Z}_j$ is the position vector in the local sequence.

The model aggregates local features in the base encoding space and the cross-modal pre-trained space, respectively:
\begin{equation}
\mathbf{r}_{T}^{B}=\operatorname{Pool}(\mathbf{H}_{T},\mathbf{q}_{T},\mathbf{m}_{T}),\quad
\mathbf{r}_{I}^{B}=\operatorname{Pool}(\mathbf{H}_{I},\mathbf{q}_{I},\mathbf{1}_{I}).
\end{equation}
Here, $\mathbf{r}_{T}^{B}$ and $\mathbf{r}_{I}^{B}$ are text- and image-side local representations in the base encoding space. $\mathbf{m}_{T}$ is the valid-position mask for the text local sequence, and $\mathbf{1}_{I}$ is an all-one mask for the image local sequence.

In the cross-modal alignment space, local aggregation is expressed as
\begin{equation}
\mathbf{r}_{T}^{C}=\operatorname{Pool}(\mathbf{P}_{T},\mathbf{q}_{T},\mathbf{m}_{C}),\quad
\mathbf{r}_{I}^{C}=\operatorname{Pool}(\mathbf{P}_{I},\mathbf{q}_{I},\mathbf{1}_{C}).
\end{equation}
Here, $\mathbf{r}_{T}^{C}$ and $\mathbf{r}_{I}^{C}$ are text- and image-side local representations in the cross-modal alignment space. $\mathbf{m}_{C}$ is the valid-position mask for the text-side cross-modal local sequence, and $\mathbf{1}_{C}$ is an all-one mask for the image-side cross-modal local sequence.

To compare local representations in the base encoding space and cross-modal pre-trained space, GAHR uses a relation operator:
\begin{equation}
\mathcal{R}(\mathbf{u},\mathbf{v})=[\mathbf{u},\mathbf{v},|\mathbf{u}-\mathbf{v}|,\mathbf{u}\odot\mathbf{v}].
\end{equation}
Here, $\mathcal{R}$ is the relation operator, $\mathbf{u}$ and $\mathbf{v}$ are the two vectors to compare, and $|\mathbf{u}-\mathbf{v}|$ denotes their elementwise absolute difference.

Based on this operator, three types of local evidence are defined uniformly as
\begin{equation}
\mathbf{e}_{T}=\phi_{eT}(\mathcal{R}(\mathbf{r}_{T}^{B},\mathbf{r}_{T}^{C})),\quad
\mathbf{e}_{I}=\phi_{eI}(\mathcal{R}(\mathbf{r}_{I}^{B},\mathbf{r}_{I}^{C})),\quad
\mathbf{e}_{TI}=\phi_{eC}(\mathcal{R}(\mathbf{r}_{T}^{C},\mathbf{r}_{I}^{C})).
\end{equation}
Here, $\mathbf{e}_{T}$, $\mathbf{e}_{I}$, and $\mathbf{e}_{TI}$ are text-side local evidence, image-side local evidence, and image--text consistency evidence; $\phi_{eT}$, $\phi_{eI}$, and $\phi_{eC}$ are their projection functions. The first two characterize differences between base and cross-modal semantics within a modality; the third characterizes local image--text consistency in the cross-modal space.

The three evidence types are concatenated to form the local representation, and the local correction head produces a correction to the global judgment:
\begin{equation}
\mathbf{h}_{L}=[\mathbf{e}_{T},\mathbf{e}_{I},\mathbf{e}_{TI}].
\end{equation}
Here, $\mathbf{h}_{L}$ is the local evidence representation.

The local correction head outputs the veracity correction:
\begin{equation}
z_{L}=f_{L}(\mathbf{h}_{L}).
\end{equation}
Here, $z_L$ is the local veracity correction and $f_L$ is the local correction mapping function.

To prevent local information from exerting a fixed-strength influence on the final result, the model learns a correction gate and adds the local correction to the global judgment as a residual term:
\begin{equation}
\gamma_{L}=\sigma(f_{\gamma}([\mathbf{h}_{G},\mathbf{h}_{L}])).
\end{equation}
Here, $\gamma_L$ is the local correction gate, and $f_{\gamma}$ is the gating mapping function.

The final veracity logit is jointly obtained from the global judgment and local correction:
\begin{equation}
z_{Y}=z_{G}+\gamma_{L}z_{L}.
\end{equation}
Here, $z_Y$ is the final veracity logit. The final prediction probability is
\begin{equation}
\hat{y}=\sigma(z_Y).
\end{equation}
Here, $\hat{y}$ is the probability that the news is predicted to be fake. This design treats the local output as a residual correction to the global judgment, thereby incorporating fine-grained evidence while preserving the overall image--text semantic judgment.

\subsection{Generativity Detection}

In news dissemination scenarios involving generative content, veracity judgment must attend not only to facts and image--text relations but also to whether text and images contain generative content. GAHR therefore includes a generativity-detection branch. While learning text-side and image-side generativity judgments, it allows generativity information to participate in representation learning for veracity judgment. The branch takes the global representation and the three types of local evidence as input:
\begin{equation}
\mathbf{h}_{A}=f_{A}([\mathbf{h}_{G},\mathbf{h}_{L}]).
\end{equation}
Here, $\mathbf{h}_{A}$ is the generativity-detection representation, and $f_A$ is the generativity-representation mapping function.

Let $a_T$ and $a_I$ be the text-side and image-side generativity labels. The branch outputs their generation probabilities:
\begin{equation}
\hat{a}_{T}=\sigma(\mathbf{w}_{T}^{A\top}\mathbf{h}_{A}+b_{T}^{A}),\quad
\hat{a}_{I}=\sigma(\mathbf{w}_{I}^{A\top}\mathbf{h}_{A}+b_{I}^{A}).
\end{equation}
Here, $\hat{a}_{T}$ and $\hat{a}_{I}$ are the probabilities that the text and image, respectively, are predicted to be generative content; $\mathbf{w}_{T}^{A}$, $\mathbf{w}_{I}^{A}$, $b_{T}^{A}$, and $b_{I}^{A}$ are parameters of the generativity classification heads.

\subsection{Optimization Objective}

The GAHR training objective consists of a veracity-classification loss and an auxiliary generativity-detection loss. For convenience, define the binary cross-entropy function as
\begin{equation}
\operatorname{BCE}(u,\hat{u})=-u\log\hat{u}-(1-u)\log(1-\hat{u}).
\end{equation}
Here, $\operatorname{BCE}$ is binary cross-entropy, $u$ is a binary label, and $\hat{u}$ is the corresponding prediction probability.

For a training batch $\mathcal{B}$, the veracity-classification loss is
\begin{equation}
\mathcal{L}_{Y}=\frac{1}{|\mathcal{B}|}\sum_{n\in\mathcal{B}}\operatorname{BCE}(y_n,\hat{y}_n).
\end{equation}
Here, $\mathcal{L}_{Y}$ is the veracity-classification loss, $|\mathcal{B}|$ is the batch size, $y_n$ is the veracity label of the $n$-th sample, and $\hat{y}_n$ is its predicted veracity probability.

For samples with generativity annotations, text- and image-generativity detection use binary cross-entropy loss. Let $\mu_n\in\{0,1\}$ indicate whether the $n$-th sample has a generativity annotation. Then,
\begin{equation}
\mathcal{L}_{G}=\frac{1}{\max\left(1,\sum_{n\in\mathcal{B}}\mu_n\right)}\sum_{n\in\mathcal{B}}\mu_n\left[\operatorname{BCE}(a_{T,n},\hat{a}_{T,n})+\operatorname{BCE}(a_{I,n},\hat{a}_{I,n})\right].
\end{equation}
Here, $\mathcal{L}_{G}$ is the generativity-detection loss; $a_{T,n}$ and $a_{I,n}$ are text- and image-generativity labels; and $\hat{a}_{T,n}$ and $\hat{a}_{I,n}$ are the corresponding predicted generation probabilities. When a batch contains no samples with generativity annotations, $\sum_{n\in\mathcal{B}}\mu_n=0$, the numerator is zero and $\mathcal{L}_{G}=0$.

The final optimization objective is
\begin{equation}
\mathcal{L}=\mathcal{L}_{Y}+\lambda_{G}\mathcal{L}_{G}.
\end{equation}
Here, $\mathcal{L}$ is the total loss and $\lambda_G$ is the weight of the auxiliary generativity-detection loss.

\section{Experiments}
\subsection{Datasets}

We conduct experiments on six multimodal fake news detection datasets: five public datasets---Weibo17 \citep{15}, Weibo21 \citep{5}, PHEME \citep{75}, GossipCop \citep{12}, and AMG \citep{78}---and the Weibo26 dataset constructed in this work. The public datasets evaluate the applicability of GAHR in conventional fake news detection scenarios, whereas Weibo26 further examines the model's veracity-assessment and generativity-recognition abilities in the presence of generative content. Dataset composition is provided in Section~\ref{sec:weibo26-analysis}. As some early datasets contain invalid links, corrupted files, images that cannot be decoded, or missing text fields, we apply the cleaning rules described in Section~\ref{sec:data-cleaning} uniformly to the public datasets. 

We distinguish the data-splitting protocol according to the collection structure of each dataset and the corresponding evaluation objective. PHEME is an event-centric rumor detection dataset in which multiple posts are collected around a limited number of predefined events. Therefore, PHEME is split at the event level, ensuring that posts associated with the same event are assigned to the same partition. For the other datasets, we use the officially released training, validation, and test partitions whenever they are available and do not re-split those datasets. For datasets without an official partition, including Weibo26, we first group samples into event-level clusters according to their semantic similarity and then divide these clusters into training, validation, and test sets with a ratio of 3:1:1. Consequently, every split constructed in this work keeps samples from the same semantic event within a single partition.

\subsection{Baselines}

We establish veracity-detection and AIGC-detection baselines according to the experimental task. The former compare the ability to distinguish real and fake news labels, whereas the latter compare the ability to recognize text-side and image-side generativity labels.

For veracity detection, we select eight representative multimodal fake news detection methods covering variational representation learning, hierarchical attention, cross-modal fusion, metric learning, neuro-symbolic reasoning, multi-scale alignment, domain-aware modeling, and mixture-of-experts architectures:
\begin{itemize}
    \item \textbf{MVAE} \citep{17} learns a joint representation through a multimodal variational autoencoder.
    \item \textbf{HMCAN} \citep{5} models multimodal relations with hierarchical contextual attention.
    \item \textbf{CAFE} \citep{80} adaptively fuses modalities through ambiguity learning.
    \item \textbf{MRML} \citep{81} enhances the boundary between real and fake news through metric learning.
    \item \textbf{NSLM} \citep{82} models deceptive patterns as latent variables and uses neuro-symbolic reasoning.
    \item \textbf{MSACA} \citep{83} combines multi-scale visual representations with cross-modal attention.
    \item \textbf{DAMMFND} \citep{84} introduces domain-aware modeling and multimodal fusion.
    \item \textbf{MIMoE-FND} \citep{85} dynamically fuses modalities with interactive mixture-of-experts.
\end{itemize}

For text-side AIGC detection, we select four representative methods covering pretrained language-model classifiers as well as approaches based on text-distribution analysis and discriminative-pattern modeling:
\begin{itemize}
    \item \textbf{BERT} \citep{59} uses a bidirectional Transformer pre-trained language model.
    \item \textbf{RoBERTa} \citep{60} adopts a more extensive pre-training strategy.
    \item \textbf{MPU} \citep{86} identifies generated content through text-distribution differences.
    \item \textbf{DP-Net} \citep{87} detects generated text through discriminative-pattern modeling.
\end{itemize}

For image-side AIGC detection, we select six representative methods covering gradient-based, frequency-domain, artifact-generalization, universal, and neighboring-pixel-based detection strategies:
\begin{itemize}
    \item \textbf{LGrad} \citep{88} identifies generated images using gradient information.
    \item \textbf{FreqNet} \citep{89} exploits frequency-domain features for generated-image detection.
    \item \textbf{SAFE} \citep{90} learns stable forged-image representations for robust detection.
    \item \textbf{UnivFD} \citep{56} performs universal detection across different generative models.
    \item \textbf{DFFreq} \citep{91} combines deep features and frequency-domain information.
    \item \textbf{NPR} \citep{92} identifies generated images through neighboring-pixel relationships.
\end{itemize}

The baselines cover multiple technical routes. To ensure fairness, all methods use the same training, validation, and test splits. For methods requiring pre-trained features, we use configurations recommended in the original papers or public implementations and report results under the same evaluation metrics.

\subsection{Implementation Details}

We apply unified preprocessing to text and images. The model uses RoBERTa \citep{60} as the text encoder, DINOv2 \citep{64} as the visual encoder, and CLIP \citep{62} as the cross-modal pre-trained encoder, projecting outputs from the different encoders into a unified 512-dimensional representation space. During training, the batch size is 64, the maximum number of epochs is 50, the optimizer is AdamW \citep{65}, the initial learning rate is $5\times10^{-5}$, and weight decay is 0.1. We use cosine annealing with warmup for learning-rate scheduling, with warmup steps accounting for 10\% of all training steps. Early stopping patience is set to 5 and the minimum improvement threshold to $10^{-4}$. To reduce the effect of random initialization, experiments are independently run with five random seeds (2026, 888, 666, 42, and 32), and mean results and standard deviations are reported.

\subsection{Veracity Detection Analysis}

Table~\ref{tab:veracity} summarizes veracity-detection results of different methods on the six datasets. To ensure a fair comparison with existing baselines, this experiment does not use generativity labels or generativity prediction scores; all methods are trained and evaluated only using veracity labels. Overall, GAHR achieves the best or near-best Accuracy and AUC on PHEME, AMG, Weibo17, Weibo21, and Weibo26, suggesting that GAHR remains competitive across datasets with different languages, scales, and sample compositions.

For class-specific metrics, GAHR performs particularly well for fake news. On PHEME, GossipCop, AMG, and Weibo26, it exhibits clear advantages in fake-news Recall or F1-score; it also achieves a high fake-news F1-score on Weibo17 and Weibo21. Some baselines perform well for real news but have insufficient fake-news recall, particularly on datasets with imbalanced class distributions. These results show that GAHR obtains relatively strong recall and F1-score for fake news on these datasets.

On Weibo26, test samples include native news and news involving generative content. Because this experiment does not introduce AI labels, the model learns the veracity-discrimination boundary only from the image--text content. GAHR achieves the best overall Accuracy, fake-news F1-score, and AUC on Weibo26, indicating that it adapts to its more complex sample composition without generativity supervision.

\clearpage

\begin{table}[pos=H]
\centering
\caption{Veracity detection results on six datasets. The results are reported as mean $\pm$ standard deviation over five runs. Red and blue indicate the best and second-best results, respectively.}
\label{tab:veracity}
\scriptsize
\setlength{\tabcolsep}{3pt}
\renewcommand{\arraystretch}{1.08}
\resizebox{\textwidth}{!}{%
\begin{tabular}{lllcccccccc}
\toprule
\multirow{2}{*}{Dataset} & \multirow{2}{*}{Method} & \multirow{2}{*}{Year} &
\multirow{2}{*}{Accuracy} & \multicolumn{3}{c}{Fake News} &
\multicolumn{3}{c}{Real News} & \multirow{2}{*}{AUC}\\
\cmidrule(lr){5-7}\cmidrule(lr){8-10}
& & & & Precision & Recall & F1-score & Precision & Recall & F1-score &\\
\midrule

\multirow[c]{9}{*}{PHEME}
& MVAE & 2019 & \msd{0.774}{0.007} & \msd{0.677}{0.007} & \msd{0.565}{0.006} & \msd{0.616}{0.006} & \msd{0.810}{0.006} & \msd{0.858}{0.006} & \msd{0.833}{0.006} & \msd{0.841}{0.007}\\
& HMCAN & 2021 & \msd{0.840}{0.014} & \msd{0.724}{0.014} & \msd{0.736}{0.014} & \msd{0.730}{0.014} & \msd{0.889}{0.015} & \msd{0.883}{0.015} & \msd{0.886}{0.014} & \msd{0.907}{0.014}\\
& CAFE & 2022 & \msd{0.816}{0.012} & \msd{0.724}{0.012} & \msd{0.606}{0.013} & \msd{0.659}{0.013} & \msd{0.847}{0.012} & \second{0.904}{0.013} & \msd{0.873}{0.013} & \msd{0.884}{0.012}\\
& MRML & 2023 & \second{0.853}{0.007} & \msd{0.722}{0.007} & \second{0.811}{0.007} & \second{0.764}{0.007} & \second{0.917}{0.008} & \msd{0.870}{0.007} & \second{0.893}{0.007} & \second{0.924}{0.006}\\
& NSLM & 2024 & \msd{0.826}{0.006} & \msd{0.715}{0.005} & \msd{0.679}{0.006} & \msd{0.697}{0.006} & \msd{0.869}{0.006} & \msd{0.887}{0.007} & \msd{0.878}{0.006} & \msd{0.897}{0.006}\\
& MSACA & 2024 & \msd{0.782}{0.012} & \msd{0.652}{0.012} & \msd{0.555}{0.012} & \msd{0.600}{0.013} & \msd{0.826}{0.012} & \msd{0.876}{0.013} & \msd{0.850}{0.012} & \msd{0.853}{0.012}\\
& DAMMFND & 2025 & \msd{0.788}{0.015} & \msd{0.642}{0.016} & \msd{0.636}{0.015} & \msd{0.640}{0.015} & \msd{0.849}{0.015} & \msd{0.852}{0.015} & \msd{0.851}{0.014} & \msd{0.860}{0.015}\\
& MIMoE-FND & 2025 & \msd{0.845}{0.014} & \best{0.771}{0.015} & \msd{0.670}{0.013} & \msd{0.716}{0.014} & \msd{0.869}{0.014} & \best{0.916}{0.014} & \msd{0.891}{0.014} & \msd{0.917}{0.014}\\
& GAHR (Ours) & 2026 & \best{0.860}{0.010} & \second{0.737}{0.010} & \best{0.818}{0.009} & \best{0.773}{0.009} & \best{0.920}{0.009} & \msd{0.876}{0.011} & \best{0.898}{0.010} & \best{0.935}{0.010}\\
\midrule

\multirow[c]{9}{*}{GossipCop}
& MVAE & 2019 & \msd{0.746}{0.014} & \msd{0.573}{0.014} & \msd{0.071}{0.014} & \msd{0.128}{0.014} & \msd{0.750}{0.014} & \msd{0.945}{0.014} & \msd{0.837}{0.015} & \msd{0.751}{0.014}\\
& HMCAN & 2021 & \msd{0.819}{0.005} & \msd{0.707}{0.006} & \msd{0.279}{0.005} & \msd{0.401}{0.004} & \msd{0.830}{0.005} & \msd{0.967}{0.005} & \msd{0.893}{0.006} & \msd{0.828}{0.005}\\
& CAFE & 2022 & \msd{0.792}{0.009} & \msd{0.624}{0.009} & \msd{0.102}{0.010} & \msd{0.176}{0.009} & \msd{0.797}{0.009} & \best{0.982}{0.009} & \msd{0.881}{0.009} & \msd{0.801}{0.009}\\
& MRML & 2023 & \msd{0.803}{0.011} & \msd{0.577}{0.012} & \msd{0.345}{0.012} & \msd{0.432}{0.013} & \msd{0.836}{0.013} & \msd{0.930}{0.011} & \msd{0.880}{0.012} & \msd{0.813}{0.012}\\
& NSLM & 2024 & \best{0.837}{0.016} & \second{0.725}{0.015} & \second{0.409}{0.015} & \second{0.523}{0.016} & \second{0.853}{0.016} & \msd{0.955}{0.016} & \best{0.902}{0.015} & \best{0.847}{0.015}\\
& MSACA & 2024 & \msd{0.810}{0.009} & \msd{0.596}{0.008} & \msd{0.372}{0.008} & \msd{0.458}{0.009} & \msd{0.843}{0.009} & \msd{0.930}{0.008} & \msd{0.884}{0.008} & \msd{0.817}{0.009}\\
& DAMMFND & 2025 & \msd{0.817}{0.014} & \msd{0.624}{0.014} & \msd{0.381}{0.013} & \msd{0.473}{0.013} & \msd{0.845}{0.013} & \msd{0.935}{0.013} & \msd{0.888}{0.013} & \msd{0.825}{0.014}\\
& MIMoE-FND & 2025 & \msd{0.825}{0.012} & \best{0.762}{0.012} & \msd{0.289}{0.012} & \msd{0.419}{0.011} & \msd{0.831}{0.013} & \second{0.973}{0.011} & \second{0.897}{0.011} & \msd{0.835}{0.012}\\
& GAHR (Ours) & 2026 & \second{0.829}{0.014} & \msd{0.603}{0.014} & \best{0.599}{0.014} & \best{0.601}{0.014} & \best{0.892}{0.013} & \msd{0.893}{0.014} & \msd{0.892}{0.014} & \second{0.840}{0.014}\\
\midrule

\multirow[c]{9}{*}{AMG}
& MVAE & 2019 & \msd{0.755}{0.007} & \msd{0.639}{0.008} & \msd{0.622}{0.009} & \msd{0.630}{0.009} & \msd{0.816}{0.008} & \msd{0.800}{0.008} & \msd{0.808}{0.008} & \msd{0.822}{0.008}\\
& HMCAN & 2021 & \msd{0.811}{0.014} & \msd{0.699}{0.014} & \msd{0.749}{0.015} & \msd{0.722}{0.015} & \msd{0.881}{0.014} & \msd{0.834}{0.014} & \msd{0.855}{0.015} & \msd{0.884}{0.014}\\
& CAFE & 2022 & \msd{0.793}{0.013} & \msd{0.682}{0.014} & \msd{0.658}{0.013} & \msd{0.670}{0.014} & \msd{0.857}{0.013} & \second{0.840}{0.013} & \msd{0.849}{0.013} & \msd{0.867}{0.013}\\
& MRML & 2023 & \msd{0.825}{0.015} & \msd{0.682}{0.016} & \second{0.820}{0.016} & \second{0.745}{0.015} & \second{0.911}{0.015} & \msd{0.819}{0.015} & \msd{0.864}{0.015} & \msd{0.901}{0.015}\\
& NSLM & 2024 & \msd{0.809}{0.006} & \msd{0.683}{0.006} & \msd{0.762}{0.006} & \msd{0.721}{0.007} & \msd{0.893}{0.006} & \msd{0.816}{0.007} & \msd{0.853}{0.006} & \msd{0.885}{0.006}\\
& MSACA & 2024 & \msd{0.784}{0.011} & \msd{0.651}{0.010} & \msd{0.668}{0.010} & \msd{0.660}{0.010} & \msd{0.855}{0.011} & \msd{0.832}{0.010} & \msd{0.844}{0.010} & \msd{0.862}{0.009}\\
& DAMMFND & 2025 & \msd{0.792}{0.005} & \msd{0.645}{0.006} & \msd{0.731}{0.006} & \msd{0.685}{0.006} & \msd{0.878}{0.006} & \msd{0.811}{0.005} & \msd{0.845}{0.005} & \msd{0.870}{0.006}\\
& MIMoE-FND & 2025 & \second{0.826}{0.009} & \best{0.738}{0.009} & \msd{0.736}{0.009} & \msd{0.738}{0.009} & \msd{0.888}{0.009} & \best{0.856}{0.009} & \second{0.871}{0.009} & \second{0.904}{0.009}\\
& GAHR (Ours) & 2026 & \best{0.843}{0.007} & \second{0.706}{0.007} & \best{0.870}{0.006} & \best{0.779}{0.006} & \best{0.932}{0.007} & \msd{0.830}{0.007} & \best{0.878}{0.007} & \best{0.925}{0.006}\\
\midrule

\multirow[c]{9}{*}{Weibo17}
& MVAE & 2019 & \msd{0.897}{0.007} & \msd{0.897}{0.008} & \msd{0.891}{0.007} & \msd{0.894}{0.008} & \msd{0.894}{0.008} & \msd{0.907}{0.008} & \msd{0.900}{0.007} & \msd{0.921}{0.007}\\
& HMCAN & 2021 & \msd{0.916}{0.006} & \msd{0.934}{0.007} & \msd{0.893}{0.007} & \msd{0.914}{0.007} & \msd{0.901}{0.008} & \msd{0.938}{0.007} & \msd{0.919}{0.006} & \msd{0.946}{0.007}\\
& CAFE & 2022 & \msd{0.930}{0.014} & \msd{0.933}{0.015} & \msd{0.924}{0.015} & \msd{0.928}{0.015} & \msd{0.928}{0.014} & \msd{0.937}{0.015} & \msd{0.931}{0.015} & \msd{0.959}{0.015}\\
& MRML & 2023 & \msd{0.937}{0.016} & \msd{0.945}{0.016} & \msd{0.924}{0.015} & \msd{0.935}{0.015} & \msd{0.930}{0.015} & \msd{0.949}{0.015} & \msd{0.939}{0.016} & \msd{0.969}{0.016}\\
& NSLM & 2024 & \msd{0.929}{0.014} & \msd{0.914}{0.014} & \msd{0.942}{0.014} & \msd{0.928}{0.014} & \msd{0.942}{0.014} & \msd{0.916}{0.014} & \msd{0.929}{0.014} & \msd{0.960}{0.013}\\
& MSACA & 2024 & \msd{0.941}{0.013} & \best{0.968}{0.014} & \msd{0.910}{0.014} & \msd{0.939}{0.014} & \msd{0.919}{0.014} & \best{0.972}{0.014} & \msd{0.944}{0.014} & \msd{0.973}{0.013}\\
& DAMMFND & 2025 & \msd{0.941}{0.008} & \second{0.957}{0.008} & \msd{0.921}{0.008} & \msd{0.939}{0.009} & \msd{0.927}{0.009} & \second{0.960}{0.008} & \msd{0.943}{0.009} & \msd{0.972}{0.009}\\
& MIMoE-FND & 2025 & \second{0.951}{0.010} & \msd{0.927}{0.010} & \best{0.976}{0.011} & \second{0.952}{0.011} & \best{0.976}{0.010} & \msd{0.924}{0.010} & \second{0.949}{0.010} & \second{0.982}{0.009}\\
& GAHR (Ours) & 2026 & \best{0.955}{0.009} & \second{0.957}{0.010} & \second{0.949}{0.009} & \best{0.953}{0.009} & \second{0.950}{0.009} & \second{0.960}{0.009} & \best{0.955}{0.010} & \best{0.988}{0.007}\\
\midrule

\multirow[c]{9}{*}{Weibo21}
& MVAE & 2019 & \msd{0.781}{0.013} & \msd{0.771}{0.014} & \msd{0.758}{0.014} & \msd{0.764}{0.014} & \msd{0.797}{0.014} & \msd{0.796}{0.014} & \msd{0.796}{0.013} & \msd{0.827}{0.013}\\
& HMCAN & 2021 & \msd{0.832}{0.015} & \msd{0.844}{0.015} & \msd{0.781}{0.015} & \msd{0.811}{0.015} & \msd{0.824}{0.015} & \msd{0.877}{0.015} & \msd{0.850}{0.015} & \msd{0.876}{0.015}\\
& CAFE & 2022 & \msd{0.821}{0.015} & \msd{0.808}{0.015} & \msd{0.802}{0.015} & \msd{0.805}{0.016} & \msd{0.832}{0.017} & \msd{0.837}{0.015} & \msd{0.836}{0.015} & \msd{0.866}{0.015}\\
& MRML & 2023 & \msd{0.872}{0.009} & \msd{0.850}{0.009} & \msd{0.878}{0.009} & \msd{0.863}{0.010} & \msd{0.893}{0.009} & \msd{0.868}{0.009} & \msd{0.881}{0.009} & \msd{0.918}{0.009}\\
& NSLM & 2024 & \msd{0.841}{0.011} & \msd{0.778}{0.012} & \msd{0.912}{0.012} & \msd{0.841}{0.012} & \msd{0.912}{0.012} & \msd{0.778}{0.012} & \msd{0.840}{0.012} & \msd{0.886}{0.012}\\
& MSACA & 2024 & \msd{0.865}{0.008} & \msd{0.868}{0.007} & \msd{0.831}{0.008} & \msd{0.850}{0.008} & \msd{0.862}{0.008} & \msd{0.893}{0.007} & \msd{0.876}{0.008} & \msd{0.910}{0.007}\\
& DAMMFND & 2025 & \second{0.902}{0.007} & \msd{0.865}{0.007} & \best{0.934}{0.008} & \second{0.897}{0.008} & \best{0.941}{0.008} & \msd{0.878}{0.008} & \msd{0.908}{0.008} & \second{0.949}{0.008}\\
& MIMoE-FND & 2025 & \msd{0.900}{0.013} & \second{0.898}{0.013} & \msd{0.878}{0.013} & \msd{0.888}{0.014} & \msd{0.902}{0.014} & \second{0.919}{0.013} & \second{0.910}{0.012} & \msd{0.946}{0.013}\\
& GAHR (Ours) & 2026 & \best{0.927}{0.009} & \best{0.923}{0.009} & \second{0.917}{0.009} & \best{0.920}{0.009} & \second{0.929}{0.009} & \best{0.935}{0.009} & \best{0.933}{0.009} & \best{0.970}{0.010}\\
\midrule

\multirow[c]{9}{*}{Weibo26}
& MVAE & 2019 & \msd{0.810}{0.005} & \msd{0.801}{0.005} & \msd{0.688}{0.005} & \msd{0.740}{0.005} & \msd{0.803}{0.006} & \msd{0.871}{0.005} & \msd{0.836}{0.005} & \msd{0.844}{0.004}\\
& HMCAN & 2021 & \msd{0.853}{0.016} & \msd{0.867}{0.016} & \msd{0.740}{0.016} & \msd{0.799}{0.015} & \msd{0.841}{0.015} & \msd{0.919}{0.016} & \msd{0.878}{0.016} & \msd{0.892}{0.016}\\
& CAFE & 2022 & \msd{0.847}{0.008} & \msd{0.845}{0.008} & \msd{0.726}{0.009} & \msd{0.781}{0.009} & \msd{0.844}{0.008} & \msd{0.906}{0.008} & \msd{0.874}{0.008} & \msd{0.886}{0.008}\\
& MRML & 2023 & \msd{0.881}{0.007} & \msd{0.864}{0.008} & \msd{0.816}{0.008} & \msd{0.839}{0.007} & \msd{0.889}{0.008} & \msd{0.913}{0.007} & \msd{0.901}{0.007} & \msd{0.922}{0.008}\\
& NSLM & 2024 & \msd{0.859}{0.016} & \msd{0.828}{0.016} & \second{0.819}{0.016} & \msd{0.823}{0.015} & \msd{0.892}{0.016} & \msd{0.865}{0.015} & \msd{0.878}{0.015} & \msd{0.900}{0.017}\\
& MSACA & 2024 & \msd{0.866}{0.005} & \msd{0.869}{0.006} & \msd{0.747}{0.006} & \msd{0.803}{0.005} & \msd{0.854}{0.006} & \second{0.933}{0.006} & \msd{0.893}{0.005} & \msd{0.908}{0.006}\\
& DAMMFND & 2025 & \msd{0.883}{0.013} & \msd{0.860}{0.014} & \msd{0.812}{0.015} & \msd{0.835}{0.015} & \second{0.897}{0.014} & \msd{0.916}{0.014} & \msd{0.905}{0.014} & \msd{0.924}{0.014}\\
& MIMoE-FND & 2025 & \second{0.895}{0.014} & \best{0.902}{0.013} & \msd{0.798}{0.014} & \second{0.847}{0.013} & \msd{0.894}{0.014} & \best{0.941}{0.013} & \best{0.917}{0.014} & \second{0.936}{0.014}\\
& GAHR (Ours) & 2026 & \best{0.910}{0.015} & \second{0.895}{0.014} & \best{0.914}{0.015} & \best{0.905}{0.015} & \best{0.924}{0.015} & \msd{0.907}{0.015} & \second{0.916}{0.015} & \best{0.959}{0.014}\\

\bottomrule
\end{tabular}}
\end{table}

\begin{table}[pos=H]
\centering
\caption{Ablation results of GAHR on AMG, Weibo21, and Weibo26. The results are reported as mean $\pm$ standard deviation over five runs. Red and blue indicate the best and second-best results, respectively.}
\label{tab:ablation}
\scriptsize
\setlength{\tabcolsep}{3pt}
\renewcommand{\arraystretch}{1.08}

\resizebox{\linewidth}{!}{%
\begin{tabular}{llcccccccc}
\toprule
\multirow{2}{*}{Dataset} & \multirow{2}{*}{Method} &
\multirow{2}{*}{Accuracy} & \multicolumn{3}{c}{Fake News} &
\multicolumn{3}{c}{Real News} & \multirow{2}{*}{AUC}\\
\cmidrule(lr){4-6}\cmidrule(lr){7-9}
& & & Precision & Recall & F1-score & Precision & Recall & F1-score &\\
\midrule

\multirow[c]{5}{*}{AMG}
& w/o Text   & \msd{0.778}{0.010} & \msd{0.622}{0.010} & \msd{0.773}{0.010} & \msd{0.689}{0.009} & \msd{0.880}{0.010} & \msd{0.781}{0.009} & \msd{0.827}{0.009} & \msd{0.830}{0.010}\\
& w/o Image  & \msd{0.809}{0.015} & \msd{0.697}{0.014} & \msd{0.797}{0.015} & \msd{0.744}{0.015} & \msd{0.873}{0.014} & \msd{0.816}{0.014} & \msd{0.842}{0.015} & \msd{0.866}{0.015}\\
& w/o Global & \second{0.817}{0.007} & \best{0.708}{0.007} & \msd{0.757}{0.007} & \msd{0.732}{0.006} & \msd{0.872}{0.007} & \best{0.845}{0.006} & \second{0.858}{0.007} & \second{0.894}{0.006}\\
& w/o Local  & \second{0.817}{0.014} & \msd{0.679}{0.014} & \second{0.844}{0.014} & \second{0.754}{0.014} & \second{0.908}{0.014} & \msd{0.806}{0.013} & \msd{0.854}{0.014} & \msd{0.891}{0.013}\\
\cmidrule(lr){2-10}
& Full Model & \best{0.843}{0.007} & \second{0.706}{0.007} & \best{0.870}{0.006} & \best{0.779}{0.006} & \best{0.932}{0.007} & \second{0.830}{0.007} & \best{0.878}{0.007} & \best{0.925}{0.006}\\

\midrule

\multirow[c]{5}{*}{Weibo21}
& w/o Text   & \msd{0.858}{0.005} & \msd{0.836}{0.006} & \msd{0.861}{0.006} & \msd{0.849}{0.006} & \msd{0.878}{0.006} & \msd{0.857}{0.006} & \msd{0.867}{0.005} & \msd{0.920}{0.006}\\
& w/o Image  & \msd{0.880}{0.013} & \msd{0.866}{0.014} & \msd{0.875}{0.014} & \msd{0.871}{0.015} & \msd{0.893}{0.015} & \msd{0.884}{0.015} & \msd{0.889}{0.015} & \msd{0.947}{0.014}\\
& w/o Global & \msd{0.886}{0.015} & \msd{0.866}{0.013} & \msd{0.892}{0.014} & \msd{0.878}{0.014} & \msd{0.904}{0.015} & \msd{0.882}{0.014} & \msd{0.894}{0.014} & \second{0.957}{0.015}\\
& w/o Local  & \second{0.901}{0.013} & \second{0.886}{0.013} & \second{0.905}{0.013} & \second{0.895}{0.013} & \second{0.915}{0.013} & \second{0.898}{0.013} & \second{0.906}{0.013} & \msd{0.951}{0.014}\\
\cmidrule(lr){2-10}
& Full Model & \best{0.927}{0.009} & \best{0.923}{0.009} & \best{0.917}{0.009} & \best{0.920}{0.009} & \best{0.929}{0.009} & \best{0.935}{0.009} & \best{0.933}{0.009} & \best{0.970}{0.010}\\

\midrule

\multirow[c]{5}{*}{Weibo26}
& w/o Text   & \msd{0.641}{0.006} & \msd{0.525}{0.005} & \msd{0.566}{0.006} & \msd{0.545}{0.006} & \msd{0.721}{0.006} & \msd{0.686}{0.005} & \msd{0.703}{0.006} & \msd{0.677}{0.005}\\
& w/o Image  & \msd{0.888}{0.006} & \msd{0.844}{0.007} & \second{0.874}{0.006} & \msd{0.859}{0.007} & \second{0.917}{0.006} & \msd{0.897}{0.007} & \msd{0.907}{0.006} & \second{0.948}{0.006}\\
& w/o Global & \msd{0.892}{0.008} & \best{0.898}{0.008} & \msd{0.817}{0.007} & \msd{0.855}{0.008} & \msd{0.889}{0.008} & \best{0.937}{0.008} & \msd{0.913}{0.008} & \msd{0.944}{0.008}\\
& w/o Local  & \second{0.897}{0.010} & \msd{0.891}{0.010} & \msd{0.842}{0.010} & \second{0.866}{0.010} & \msd{0.900}{0.010} & \second{0.930}{0.010} & \second{0.915}{0.010} & \msd{0.933}{0.010}\\
\cmidrule(lr){2-10}
& Full Model & \best{0.910}{0.015} & \second{0.895}{0.014} & \best{0.914}{0.015} & \best{0.905}{0.015} & \best{0.924}{0.015} & \msd{0.907}{0.015} & \best{0.916}{0.015} & \best{0.959}{0.014}\\

\bottomrule
\end{tabular}%
}
\end{table}

\subsection{Ablation Study}

Table~\ref{tab:ablation} reports ablation results for GAHR on AMG, Weibo21, and Weibo26. We separately remove the text modality, image modality, global judgment, and local correction to examine the effect of information sources and reasoning levels. The full model achieves the best overall performance on all three datasets, indicating that the complete combination of components is beneficial under these settings.

Removing the text modality causes a marked performance decline on all datasets, indicating that text still plays the primary discriminative role. Removing images also weakens performance, showing that visual information provides supplementary evidence. Removing either global judgment or local correction also lowers performance. In particular, on Weibo26, the full model outperforms both structural-ablation variants in fake-news Recall and F1-score, indicating complementary roles for global judgment and local correction in samples involving generative content.

\subsection{AIGC Detection Analysis}

Weibo26 provides separate generativity labels for the text and image modalities. The generative construction of Weibo26 is single-sided: ICRN and ICFN contain generated text paired with native images, whereas TCRN and TCFN contain generated images paired with native text. Based on this construction, we evaluate text-side and image-side AIGC detection as two separate unimodal tasks.

To ensure a fair comparison with unimodal AIGC baselines, we apply modality-specific ablations to GAHR, although the full GAHR framework uses a multimodal generativity branch. For text-side AIGC detection, the image input, visual representations, image-dependent local evidence, and image--text interaction components are removed, and GAHR makes predictions using only text-side representations. For image-side AIGC detection, the text input, textual representations, text-dependent local evidence, and image--text interaction components are removed, and GAHR makes predictions using only image-side representations. 

Table~\ref{tab:aigc-text} and Table~\ref{tab:aigc-image} present the text-side and image-side generativity-detection results on Weibo26, respectively. The text-side and image-side results are reported using two modality-specific variants, denoted as GAHR-T and GAHR-I, respectively. GAHR achieves the best overall performance under both modality-specific settings.

\begin{table}[pos=H]
\centering
\caption{Unimodal text-side AIGC detection results on Weibo26. Red and blue indicate the best and second-best results, respectively.}
\label{tab:aigc-text}
\footnotesize
\setlength{\tabcolsep}{3pt}
\renewcommand{\arraystretch}{1.08}

\begin{tabular*}{\linewidth}{@{\extracolsep{\fill}}lcccccccc@{}}
\toprule
\multirow{2}{*}{Method} & \multirow{2}{*}{Accuracy} &
\multicolumn{3}{c}{Generated Content} &
\multicolumn{3}{c}{Native Content} &
\multirow{2}{*}{AUC}\\
\cmidrule(lr){3-5}\cmidrule(lr){6-8}
& & Precision & Recall & F1-score &
Precision & Recall & F1-score &\\
\midrule

BERT
& \msd{0.770}{0.013}
& \msd{0.681}{0.012}
& \msd{0.638}{0.013}
& \msd{0.659}{0.012}
& \msd{0.810}{0.013}
& \msd{0.838}{0.013}
& \msd{0.824}{0.013}
& \msd{0.844}{0.013}\\

RoBERTa
& \msd{0.780}{0.008}
& \msd{0.696}{0.007}
& \msd{0.655}{0.007}
& \msd{0.674}{0.008}
& \msd{0.820}{0.007}
& \msd{0.845}{0.008}
& \msd{0.832}{0.007}
& \msd{0.855}{0.008}\\

MPU
& \msd{0.800}{0.014}
& \msd{0.727}{0.014}
& \msd{0.678}{0.014}
& \msd{0.702}{0.014}
& \msd{0.832}{0.015}
& \msd{0.863}{0.015}
& \msd{0.847}{0.015}
& \msd{0.877}{0.015}\\

DP-Net
& \second{0.810}{0.013}
& \second{0.746}{0.013}
& \second{0.688}{0.013}
& \second{0.716}{0.013}
& \second{0.838}{0.013}
& \second{0.873}{0.014}
& \second{0.855}{0.014}
& \second{0.890}{0.013}\\

\cmidrule(lr){1-9}

GAHR-T (Ours)
& \best{0.821}{0.013}
& \best{0.766}{0.013}
& \best{0.704}{0.012}
& \best{0.733}{0.013}
& \best{0.848}{0.013}
& \best{0.883}{0.013}
& \best{0.865}{0.014}
& \best{0.908}{0.012}\\

\bottomrule
\end{tabular*}
\end{table}

\begin{table}[pos=H]
\centering
\caption{Unimodal image-side AIGC detection results on Weibo26. Red and blue indicate the best and second-best results, respectively.}
\label{tab:aigc-image}
\footnotesize
\setlength{\tabcolsep}{3pt}
\renewcommand{\arraystretch}{1.08}

\begin{tabular*}{\linewidth}{@{\extracolsep{\fill}}lcccccccc@{}}
\toprule
\multirow{2}{*}{Method} & \multirow{2}{*}{Accuracy} &
\multicolumn{3}{c}{Generated Content} &
\multicolumn{3}{c}{Native Content} &
\multirow{2}{*}{AUC}\\
\cmidrule(lr){3-5}\cmidrule(lr){6-8}
& & Precision & Recall & F1-score &
Precision & Recall & F1-score &\\
\midrule

LGrad
& \msd{0.822}{0.006}
& \msd{0.676}{0.006}
& \msd{0.759}{0.006}
& \msd{0.715}{0.006}
& \msd{0.892}{0.006}
& \msd{0.845}{0.006}
& \msd{0.867}{0.006}
& \msd{0.866}{0.005}\\

FreqNet
& \msd{0.833}{0.011}
& \msd{0.692}{0.012}
& \msd{0.773}{0.011}
& \msd{0.730}{0.011}
& \msd{0.897}{0.011}
& \msd{0.851}{0.012}
& \msd{0.875}{0.011}
& \msd{0.876}{0.011}\\

SAFE
& \msd{0.882}{0.007}
& \msd{0.776}{0.007}
& \msd{0.839}{0.007}
& \msd{0.806}{0.006}
& \msd{0.928}{0.006}
& \msd{0.896}{0.006}
& \msd{0.912}{0.007}
& \msd{0.924}{0.006}\\

UnivFD
& \msd{0.935}{0.015}
& \msd{0.860}{0.014}
& \msd{0.908}{0.015}
& \msd{0.884}{0.015}
& \msd{0.958}{0.014}
& \msd{0.937}{0.016}
& \msd{0.947}{0.014}
& \msd{0.966}{0.015}\\

DFFreq
& \msd{0.922}{0.010}
& \msd{0.841}{0.010}
& \msd{0.899}{0.010}
& \msd{0.869}{0.010}
& \msd{0.955}{0.010}
& \msd{0.927}{0.010}
& \msd{0.941}{0.010}
& \msd{0.957}{0.010}\\

NPR
& \second{0.944}{0.013}
& \second{0.878}{0.014}
& \second{0.923}{0.014}
& \second{0.900}{0.013}
& \second{0.965}{0.014}
& \second{0.944}{0.014}
& \second{0.955}{0.013}
& \second{0.973}{0.014}\\

\cmidrule(lr){1-9}

GAHR-I (Ours)
& \best{0.994}{0.004}
& \best{0.985}{0.005}
& \best{0.992}{0.004}
& \best{0.988}{0.005}
& \best{0.996}{0.003}
& \best{0.994}{0.004}
& \best{0.995}{0.004}
& \best{0.999}{0.001}\\

\bottomrule
\end{tabular*}
\end{table}

\begin{figure}[pos=H]
  \centering
  \includegraphics[width=\textwidth]{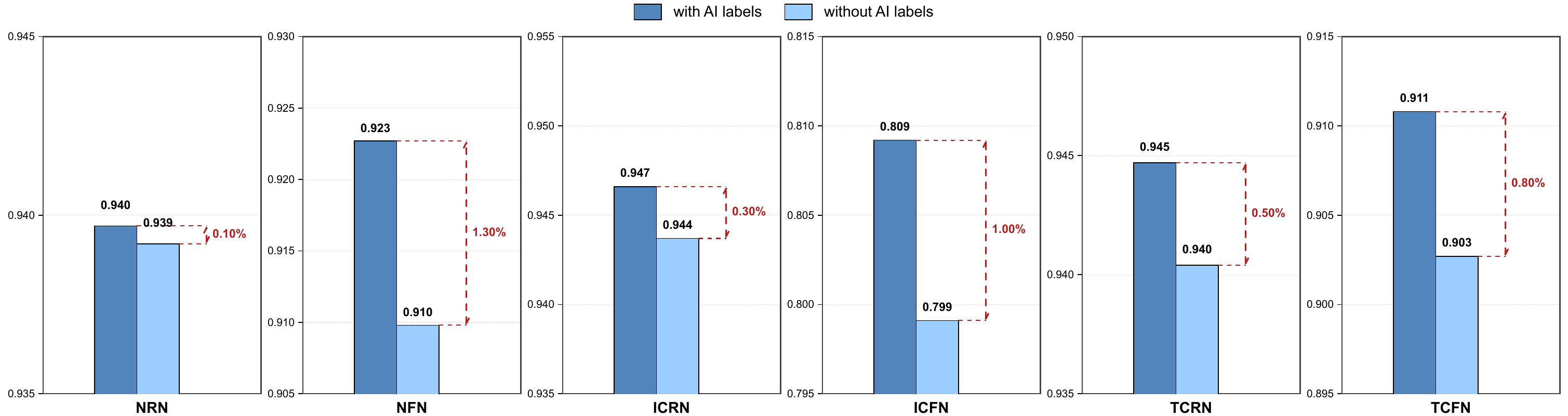}
  \caption{Veracity detection with and without generativity labels across the six Weibo26 sample types.}
  \label{fig:generativity-supervision}
\end{figure}

Figure~\ref{fig:generativity-supervision} compares veracity-detection results with and without generativity labels. Accuracy improves for all six sample types when generativity labels are used, with more pronounced gains for the fake-news types NFN, ICFN, and TCFN. Generativity labels thus provide effective information related to sample construction, especially improving fake-news recognition.

\begin{figure}[pos=htbp]
  \centering
  \includegraphics[width=\textwidth]{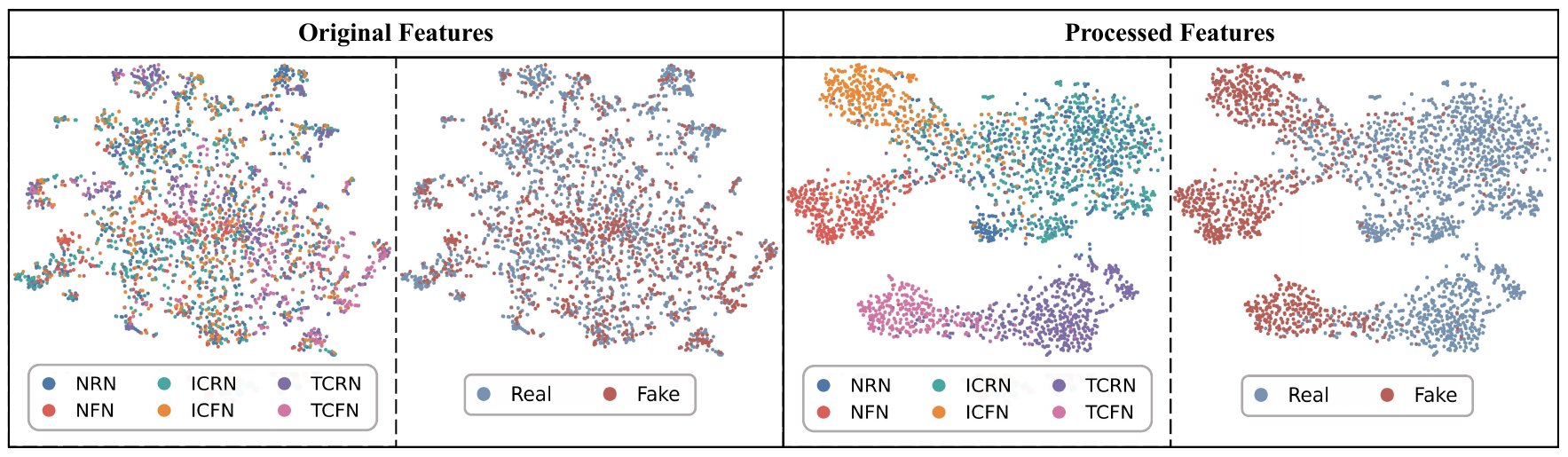}
  \caption{t-SNE visualization of generativity-related representations before and after GAHR processing.}
  \label{fig:tsne}
\end{figure}

Figure~\ref{fig:tsne} presents a t-SNE visualization of generativity-related representations. The left side is the original feature space and the right side is the feature space after GAHR processing. The six sample types are more entangled in the original features, and real and fake samples overlap substantially. After GAHR modeling, the six sample types appear more separated in the two-dimensional visualization, and the real and fake samples show less overlap. This visualization suggests that GAHR organizes the representations according to generative involvement and veracity-related information.

\subsection{Parameter Analysis}

\begin{figure}[pos=htbp]
  \centering
  \includegraphics[width=\textwidth]{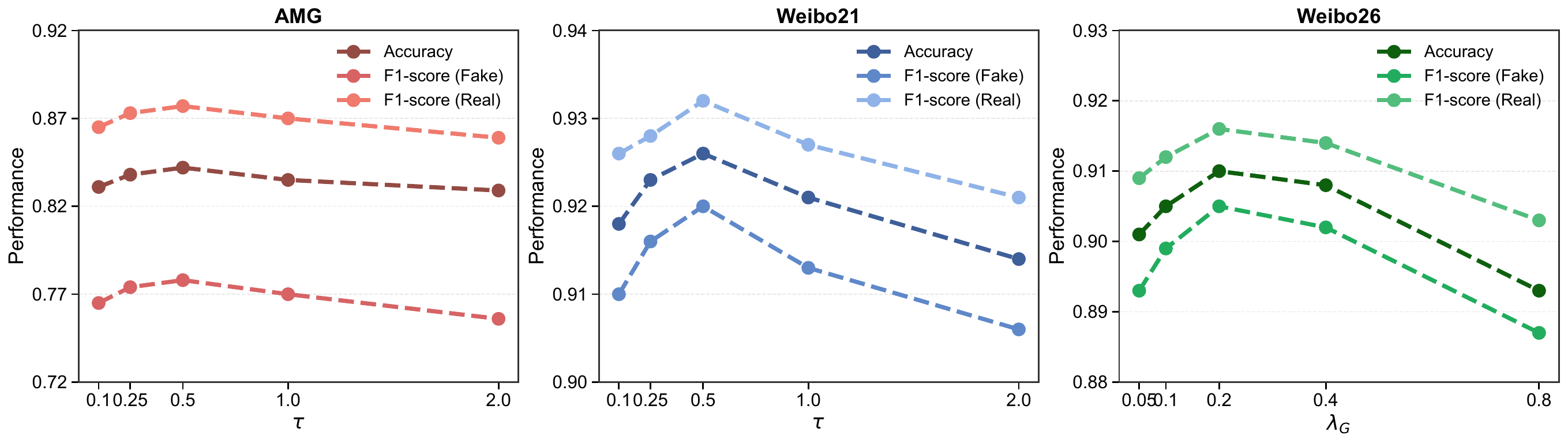}
  \caption{Effects of the attention temperature $\tau$ and the generativity-detection loss weight $\lambda_G$ on model performance.}
  \label{fig:parameter-analysis}
\end{figure}

The parameter analysis examines the influence of the attention temperature $\tau$ and the generativity-detection loss weight $\lambda_G$. Figure~\ref{fig:parameter-analysis} presents the parameter-varying results on AMG, Weibo21, and Weibo26. AMG and Weibo21 are used to analyze $\tau$ for veracity detection, whereas Weibo26 is used to analyze $\lambda_G$ after generativity supervision is introduced.

At intermediate temperatures, the model achieves better Accuracy, fake-news F1-score, and real-news F1-score. The evaluated metrics decline when $\tau$ is either too small or too large. This pattern is consistent with overly concentrated attention at low temperatures and overly diffuse attention at high temperatures. When the generativity-detection loss weight is small, supervision contributes little to veracity detection; performance then improves as it increases, peaks within a moderate range, and falls when it grows further. Generativity information can enhance veracity judgment, but its training weight must be balanced with the main task.

\subsection{Case Study}

\begin{figure}[pos=htbp]
  \centering
  \includegraphics[width=0.92\textwidth]{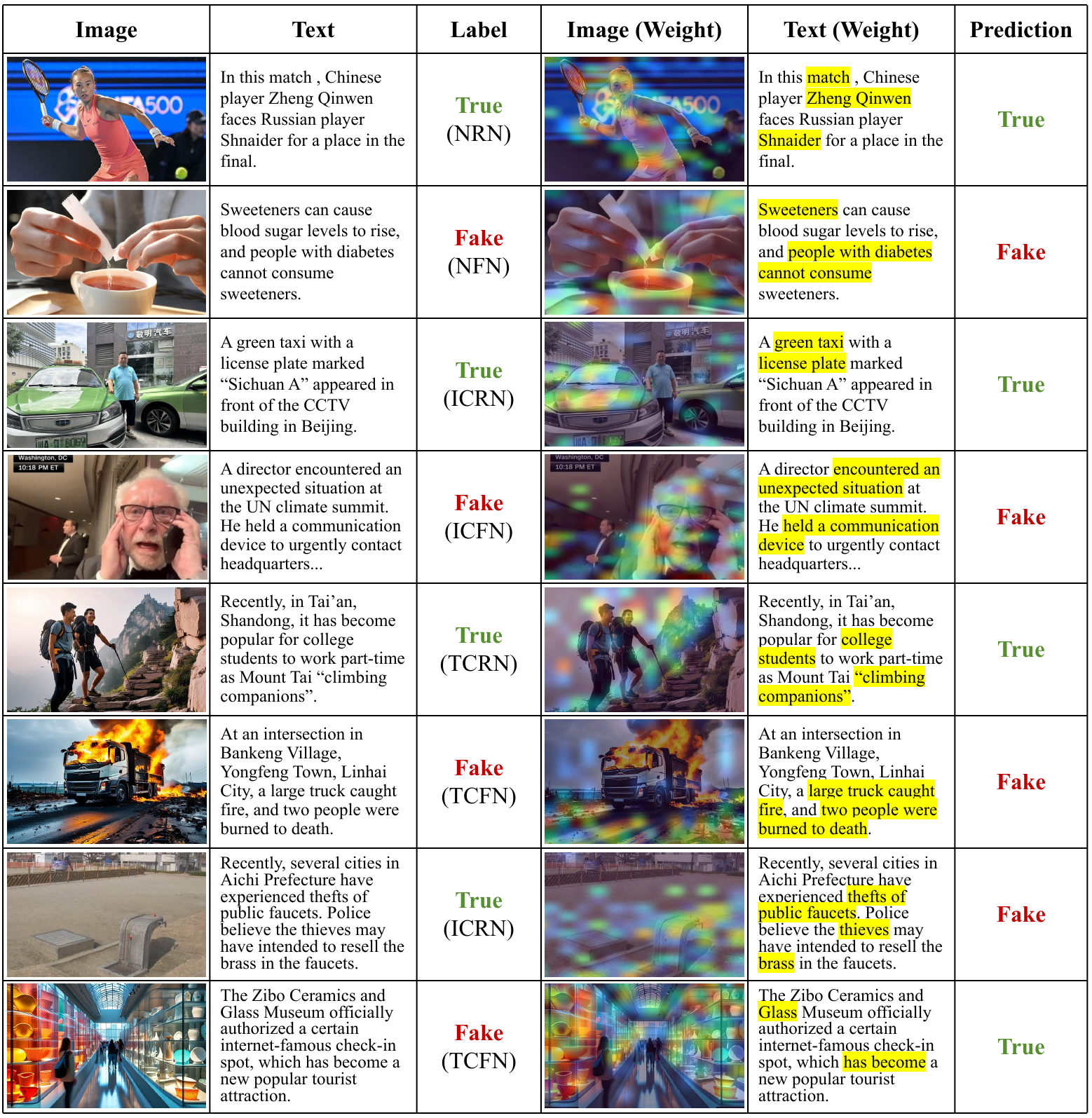}
  \caption{Representative cases under different types of generative involvement.}
  \label{fig:cases}
\end{figure}

Figure~\ref{fig:cases} presents representative cases from different types of generative involvement, together with the model predictions and highlighted textual and visual regions. In the correctly classified cases, the highlighted evidence includes information related to the news subjects, entities, locations, event descriptions, and displayed objects. For real-news samples, the highlighted regions are mainly associated with the reported event and its participants, whereas for fake-news samples, the model focuses on textual or visual elements related to the corresponding narrative. 

The figure also includes two misclassified cases. For the ICRN sample, the model attends to fact-related fragments such as ``public faucets'' ``thefts'' and ``brass''. Because ICRN represents factually consistent generated text,  it fails to recognize that the image--text pair is visually coherent but grounded in a false narrative. For the TCFN sample, the model focuses on isolated words such as ``Glass'' and ``become'' as well as display-related visual regions. However, it fails to recognize that the image--text pair is visually coherent but grounded in a false narrative, and consequently classifies the sample as real. These errors indicate the difficulty of distinguishing factual veracity from surface-level linguistic or cross-modal coherence.

\section{Conclusion}

This work investigates multimodal fake news detection when generative content participates in dissemination. To address the limited characterization of generativity information in existing datasets, we construct Weibo26. Based on social-media news image--text pairs, it contains the six types NRN, NFN, ICRN, ICFN, TCRN, and TCFN and provides both veracity and generativity labels, providing a dataset for jointly evaluating news veracity and generative-content detection after generative content participates in dissemination. We propose the Generativity-Aware Hierarchical Reasoning (GAHR) framework, which incorporates global judgment, local correction, and generativity detection into a unified model. GAHR first models the overall relationship between news text and images, then uses local evidence to correct the global judgment, and learns text-side and image-side generativity judgments on data with annotations. Experimental results show that GAHR achieves competitive veracity-detection performance on the evaluated datasets and obtains strong results on the generative-content detection tasks of Weibo26.

\bibliographystyle{cas-model2-names}
\bibliography{reference}
\end{document}